\documentclass[lettersize,journal]{IEEEtran}

\usepackage{amsmath}
\usepackage{amssymb}
\usepackage{booktabs}
\usepackage[font=small,skip=2pt]{caption}
\usepackage{comment}
\usepackage{graphicx}
\usepackage{multirow}
\usepackage{adjustbox}
\usepackage{pifont}
\usepackage{cite}
\usepackage[table]{xcolor}
\usepackage{graphicx} 
\usepackage{booktabs} 
\usepackage{makecell} 

\makeatletter
\let\NAT@parse\undefined
\makeatother

\definecolor{darkgreen}{RGB}{0, 150, 0}
\definecolor{darkred}{RGB}{200, 0, 0}
\definecolor{darkblue}{RGB}{0, 0, 200}
\definecolor{cvprblue}{rgb}{0.21,0.49,0.74}

\definecolor{navy}{RGB}{0, 0, 128}
\definecolor{emerald}{RGB}{0, 128, 0}
\definecolor{crimson}{RGB}{220, 20, 60}
\definecolor{golden}{RGB}{255, 215, 0}
\definecolor{magenta}{RGB}{255, 0, 255}
\definecolor{cyanblue}{RGB}{0, 255, 255}

\newcommand{\ch}{{\color{darkgreen} \ding{51}}}
\newcommand{\xm}{{\color{darkred} \ding{55}}}

\definecolor{gray9}{gray}{.9}
\definecolor{gray95}{gray}{.95}
\definecolor{gray8}{gray}{.8}
\definecolor{gray85}{gray}{.85}

\usepackage[colorlinks,pagebackref=true,citecolor=cvprblue,bookmarks=false,hypertexnames=true]{hyperref} 
\newcommand{\sota}{state-of-the-art}
\newcommand{\algoname}{DaF-BEVSeg}

\begin{document}
\title{\LARGE \bf
\algoname: Distortion-aware Fisheye Camera based \\ Bird's Eye View Segmentation with Occlusion Reasoning}

\author{Senthil Yogamani$^{1}$, David Unger$^{2}$, Venkatraman Narayanan$^{3}$ and Varun Ravi Kumar$^{3}$\\
$^1$Automated Driving, QT Technologies Ireland Limited\\
$^2$Internship at Qualcomm Technologies International GmbH\\
$^3$Automated Driving, Qualcomm Technologies, Inc.\\
}

\maketitle
\thispagestyle{empty}
\pagestyle{empty}
\begin{abstract}

Semantic segmentation is an effective way to perform scene understanding. Recently, segmentation in 3D  Bird's Eye View (BEV) space has become popular as its directly used by drive policy. However, there is limited work on BEV segmentation for surround-view fisheye cameras, commonly used in commercial vehicles. As this task has no real-world public dataset and existing synthetic datasets do not handle amodal regions due to occlusion, we create a synthetic dataset using the Cognata simulator comprising diverse road types, weather, and lighting conditions. We generalize the BEV segmentation to work with any camera model; this is useful for mixing diverse cameras. 
We implement a baseline by applying cylindrical rectification on the fisheye images and using a standard LSS-based BEV segmentation model. We demonstrate that we can achieve better performance without undistortion, which has the adverse effects of increased runtime due to pre-processing, reduced field-of-view, and resampling artifacts. Further, we introduce a distortion-aware learnable BEV pooling strategy that is more effective for the fisheye cameras.  We extend the model with an occlusion reasoning module, which is critical for estimating in BEV space. Qualitative performance of~\algoname~is showcased in the video at \url{https://streamable.com/ge4v51}.

\end{abstract}
\section{Introduction}

\label{sec:intro}
\begin{figure}[!t]
  \newcommand{\turnwidth}{0.485\columnwidth}

\newcommand{\imlabel}[2]{\includegraphics[width=0.49\columnwidth]{#1}%
\raisebox{2pt}{\makebox[-2pt][r]{\footnotesize #2}}}

\begin{tabular}{@{\hskip 0mm}c@{\hskip 1mm}c}
\centering
\imlabel{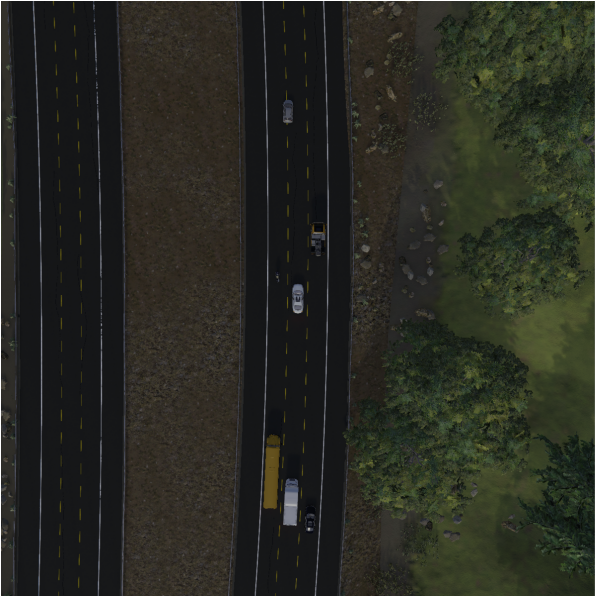}
{} &
\imlabel{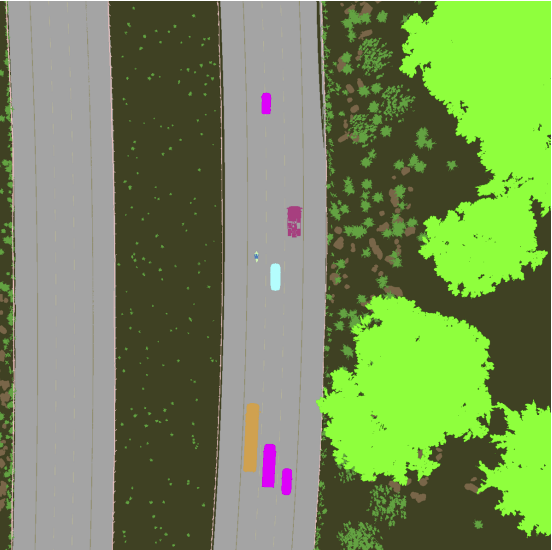}
{} \\
\end{tabular}
  \caption{Semantic encoding of the BEV space. Left: BEV camera image. Right: Scene's semantic representation.}
  \label{fig:bev_space}
\end{figure}

In the pursuit of advancing Automated Driving (AD) and Robotics, Bird's Eye View (BEV) grid generation has emerged as a critical task, aiming to provide a comprehensive understanding of the surrounding environment. As such, several \sota ~works on AD tasks like 3D Object Detection~\cite{huang2021bevdet, rashed2021bev, liu2023sparsebev, Li2022BEVFormerLB, Philion2020LiftSS, Zhang_2023_ICCV, Klingner2023X3KDKD}, Semantic Segmentation~\cite{Borse2022XAlignCC, Li2023BEVDGCL, Peng2022BEVSegFormerBE, Pan2023BAEFormerBA}, Multi-task learning~\cite{Liu2022BEVFusionMM, Huang2023FastBEVTR, Xie2022M2BEVMJ, Wang2023UniTRAU, Pham2023NVAutoNetFA}, operates on a BEV grid. Semantic BEV segmentation is the task of creating a 2D grid that rasterizes the real
world into equal-sized grid cells that each contain semantic content about the object present at the corresponding real-world location. An example of what the semantic BEV grid looks like can be seen in Figure \ref{fig:bev_space}.\par

Enabling safe automated driving requires a diverse sensor
set containing many different cameras. Given the large Field of View (FOV) of the fisheye (FE) Cameras, they are quickly becoming ubiquitous in the AD sensor setup. However, academic research on automated driving focuses heavily on pinhole cameras because
of their predominance in open-source automated driving datasets \cite{fong2021panoptic, Sun_2020_CVPR}. Furthermore, fisheye cameras introduce more substantial distortions to the images. Hence, fisheye cameras require a different camera projection model, and it is non-trivial to extend standard \sota ~works on pinhole cameras to fisheye cameras.\par

Four wide angle fisheye cameras form the basic sensor set in automotive systems for near-field sensing in combination with Ultrasonics sensor ~\cite{kumar2023surround, popperli2019capsule}. The limited literature in various fisheye perception tasks such as object detection~\cite{rashed2020fisheyeyolo,  yahiaoui2019overview}, depth estimation~\cite{kumar2021svdistnet, kumar2018near} , localization~\cite{tripathi2020trained, Konrad2021FisheyeSuperPointKD} , motion estimation~\cite{shen2023optical, mohamed2021monocular, rashed2019motion}, semantic segmentation~\cite{ramachandran2021woodscape, rashed2019optical}, blockage and adverse weather detection~\cite{uricar2019desoiling, dhananjaya2021weather}, multi-task learning~\cite{sistu2019real, sistu2019neurall, leang2020dynamic} and near-field perception systems~\cite{kumar2021omnidet, eising2021near, kumar2023surround, uricar2019challenges} indicate that special attention and design is needed to handle the large radial distortion.

The work on Lift-Splat-Shoot (LSS)~\cite{Philion2020LiftSS} forms the baseline for generating a BEV map from standard pinhole cameras. Many \sota ~works adapt LSS for downstream AD tasks (\cite{Borse2022XAlignCC, Liu2022BEVFusionMM, huang2021bevdet, mohapatra2021bevdetnet}). While pinhole camera-based BEV map generation and multi-modal sensor fusion have been studied extensively, they still need to be explored for fisheye cameras.

This work proposes a novel approach to extend LSS to fisheye cameras. It further extends our capability to perform multi-sensor fusion in BEV space with fisheye cameras. In addition, we propose a \sota ~ architecture that performs Semantic Segmentation in BEV Space for fisheye cameras. We incorporated a multi-task head that makes classification and occlusion predictions for each cell instead of a simple pixel-wise classification. It enables us to circumvent network hallucinations in occluded regions, which is predominant in fisheye camera images due to other non-linear optical characteristics. Further, we also developed an efficient pooling mechanism to improve the BEV grid generation from multiple sensors that cater to the fisheye lens characteristics. We also perform an extensive evaluation of our architecture on synthetic datasets.\par
\noindent Our main contributions include:
\begin{itemize}
\item Creation of a fisheye BEV segmentation dataset with occlusion masks using a commercial-grade simulator.
\item Design of a novel distortion-aware learnable pooling strategy using camera intrinsics for adaptation.
\item Generalized framework for BEV semantic segmentation from raw images supporting various camera models.
\item A end-to-end multi-task model that provides semantic classes and occlusion reasoning in ambiguous scenarios.
\end{itemize}
\section{Related Work}
\label{sec:related}

\subsection{BEV Generation for Automated Driving Tasks}

Recent literature explores BEV generation from camera images in 3 broad approaches. The first category of BEV transformation architectures involves Multilayer Perceptron (MLP)-based methods, where the network learns the transformation from image to BEV space within its connections ~\cite{lu2019monocular, pan2020cross, roddick2020predicting}. Attention-based methods use cross-attention, with queries in the BEV space for keys and values in the image space. TIIM~\cite{saha2022translating} leverages the symmetry between an image column and a ray in BEV space, while CVT~\cite{zhou2022cross} treats the entire image-to-BEV transformation as a translation problem. While attention-based approaches have shown promise, they are computationally expensive and may not be suitable for fisheye cameras due to specific transformations and limited computational constraints.

Lastly, geometry-based methods utilize camera parameters such as position, orientation, focal length, and sensor-shift. Inverse Perspective Mapping (IPM)~\cite{mallot1991inverse, kim2019deep} assumes a flat ground surface, projecting each pixel into BEV space. While simpler, this approach introduces distortions, making it less suitable for real-world scenarios. Some methods~\cite{zhu2018generative} use generative methods to reduce the distortions. Cam2BEV~\cite{reiher2020sim2real} employs IPM but compensates for errors through learning in the BEV encoder. Another method is to use depth estimates for projecting image features accurately into 3D space~\cite{Philion2020LiftSS, huang2021bevdet}.
\subsection{Semantic Segmentation on Generic Camera}

As mentioned in Section \ref{sec:intro}, standard vision pipelines apply only for pinhole cameras and do not scale well for generic camera applications such as fisheye cameras. NVAutoNet~\cite{Pham2023NVAutoNetFA} proposes a framework to process generic camera images for automated driving perception applications using an MLP-based BEV generation. However, semantic segmentation is not considered in the approach. F2BEV~\cite{samani2023f2bev} proposes a distortion-aware spatial cross-attention, using normalized undistorted images in conjunction with distorted images. FisheyePixPro~\cite{ramchandra2022fisheyepixpro} proposes a 2D self-supervised semantic segmentation based on contrastive learning. Other fisheye camera-specific approaches propose overlapping pyramid pooling ~\cite{deng2017cnn} and deformable convolution~\cite{deng2019restricted} for semantic segmentation. Dense depth/structure estimation methods on fisheye cameras, like~\cite{hane2014real, won2019omnimvs} utilize specific camera projection models or use camera parameters as input~\cite{kumar2020fisheyedistancenet}
\section{Method}
\label{sec:method}
This section presents our approach to learning bird's-eye-view representations of scenes from image data captured by a surround fisheye camera rig. Our goal is, given \(n\) surround-view fisheye images \({X_k \in R^{3 \times H \times W}}\) each with an extrinsic matrix \(E_k \in R^{3 \times 4}\) and an intrinsic matrix \(I_k \in R^{3 \times 3}\), and we seek to find a rasterized representation of the scene in the BEV coordinate frame \(y \in R^{C \times X \times Y}\) with semantic information for each cell. The end-to-end architecture of~\algoname is outlined in Figure \ref{fig:architecture}.\par

\subsection{Fisheye Camera BEV features}

First, similar to the LSS \cite{Philion2020LiftSS} approach, we use the camera parameters to project the image features from the image encoder into the 3D world. The camera parameters map the image feature to the corresponding ray in the BEV map, while the estimated depth determines the distance. 
For this to work in Fisheye cameras, we convert the camera features into a direction vector for each feature projected into 3D space. The camera direction vector is moved to the vehicle coordinate system using projection geometry. Firstly, the camera features in pixel coordinates, {\(u, v\)}, are converted to camera co-ordinates, {\(x, y\)}, using the camera intrinsic matrix (\(K\)). As depicted in Equation \ref{eqn:kbe}, using the Kannala-Brandt model \cite{1642666}, we generate spherical angles (\(\theta, \varphi\)) from the camera coordinate features.
\begin{equation} \label{eqn:kbe}
    \begin{split}
        r = \sqrt{x^2 + y^2} \\
        \theta = p_1.r + p_2.r^2 + p_3.r^3 +...+p_9.r^9 \\
        \varphi = \arctan\big(\frac{y}{x}\big)
    \end{split}
\end{equation}
We use a generic equation for radial distortion angle (\(\theta\)) here. Our method generalizes the inverse mappings of radial distortion models, such as \textit{Polynomial, UCM, eUCM, Rectilinear, Stereographic, Double Sphere}.
based on the derivation by  \cite{kumar2020unrectdepthnet}.

\begin{figure}[!t]
  \begin{tabular}{@{\hskip 0mm}c@{\hskip 1mm}c}
\centering
\includegraphics[width=0.51\columnwidth]{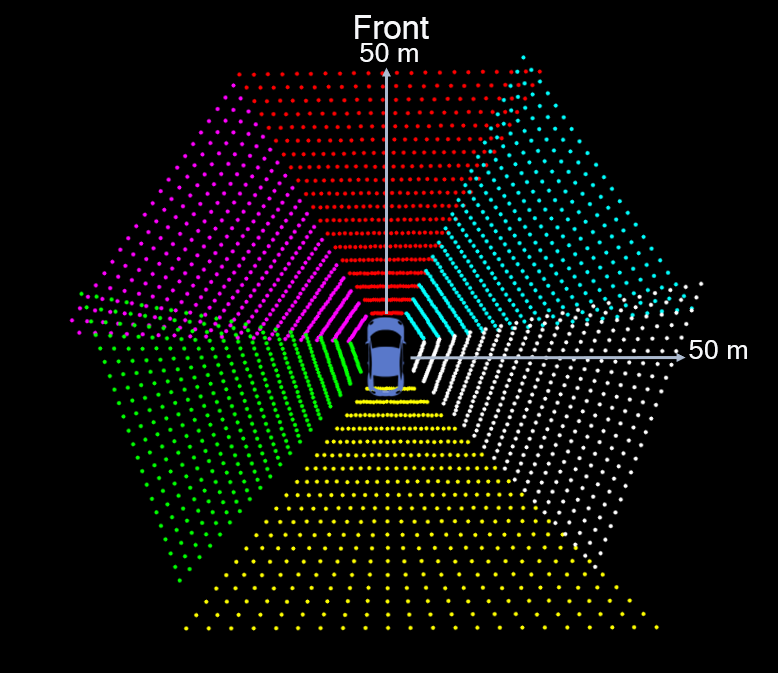}
\raisebox{2pt}{\makebox[-5pt][r]{\footnotesize  \hspace{-0.98\columnwidth}\textcolor{white}{NuScenes - Pinhole}}} &
\includegraphics[width=0.46\columnwidth]{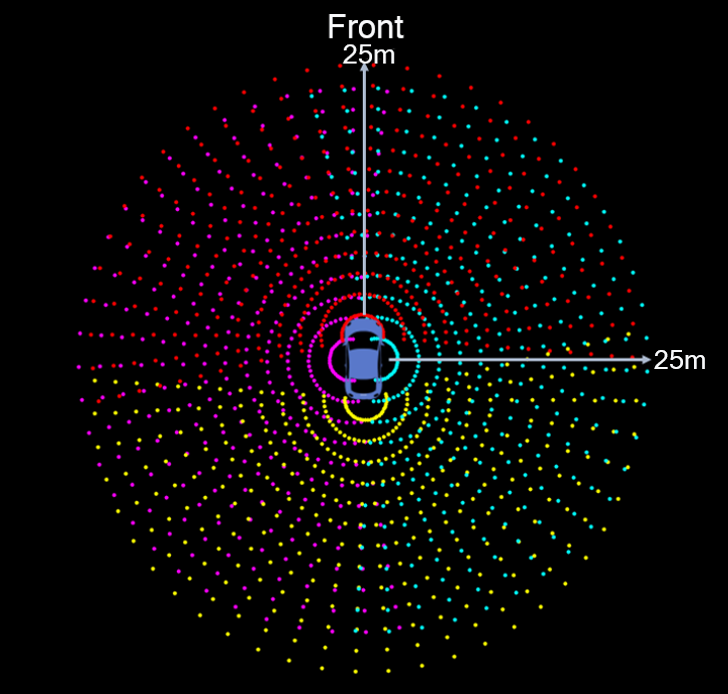}
\raisebox{2pt}{\makebox[-5pt][r]{\footnotesize \hspace{-0.98\columnwidth}\textcolor{white}{Cognata - Fisheye}}} \\
\end{tabular}

  \caption{\textbf{The visulization of BEV Grid space from each camera pixel}. \footnotesize{On the \textbf{Left} is camera BEV grid sample from NuScenes dataset~\cite{caesar2020nuscenes} color-coded as \textcolor{crimson}{FRONT}, \textcolor{cyanblue}{FRONT-RIGHT}, \textcolor{lightgray}{BACK-RIGHT}, \textcolor{yellow}{BACK}, \textcolor{green}{BACK-LEFT}, \textcolor{magenta}{FRONT-LEFT}. On the \textbf{Right} is camera BEV grid sample from our Cognata dataset color-coded as \textcolor{crimson}{FRONT}, \textcolor{green}{RIGHT}, \textcolor{yellow}{BACK}, \textcolor{magenta}{LEFT}}.}
  \label{fig:lss_feats}
\end{figure}

Our method supports the inverse mappings of radial distortion models summarized below:

\begin{enumerate}
\item Polynomial: \(r = a_1 \theta + a_2 \theta^2 + a_3 \theta^3 + a_4 \theta^4\)
\item UCM: \(r = f\cdot\sin\theta / (\cos\theta + \xi)\)
\item eUCM: \(r = f\cdot\frac{\sin\theta}{\cos\theta + \alpha\left(\sqrt{\beta\cdot \sin^2\theta + \cos^2\theta} - \cos\theta\right)}\)
\item Rectilinear: \(r = f \cdot \tan\theta\)
\item Stereographic: \(r = 2 f \cdot \tan(\theta/2)\) 
\item Double Sphere:\newline \(r = f\cdot \frac{\sin\theta}{\alpha\sqrt{\sin^2\theta + (\xi + \cos\theta)^2} + (1-\alpha)(\xi + \cos\theta)}\)
\end{enumerate} 
Where \(f\) represents the focal length of the camera. The Rectilinear and Stereographic models are unsuitable for fisheye cameras but are added for completeness, whereas the Polynomial model is the most used. The spherical angles (\(\theta, \varphi\)) are used to generate the direction vectors (\(X, Y, Z\)), following the equation \ref{eqn:dirvec}. The direction vectors are multiplied with depth estimates in a subsequent step to obtain the final 3D vectors. 
\begin{gather} \label{eqn:dirvec}
    \begin{pmatrix} X \\ Y \\ Z \end{pmatrix} = r \begin{pmatrix}
        \cos{\varphi}.\sin{\theta} \\
        \sin{\varphi}.\sin{\theta} \\
        \cos{\theta}
    \end{pmatrix}
\end{gather}
The BEV grid is generated by \textit{Splatting}~\cite{Philion2020LiftSS} the depth estimates with the direction vectors, (\(X, Y, Z\)). Unlike traditional cameras with a narrower field of view, fisheye lenses capture a wide-angle view, resulting in significant overlap between adjacent images. This overlap means that features from multiple camera perspectives contribute to the same areas in the BEV grid, as seen in Figure~\ref{fig:lss_feats}, creating a rich and redundant source of information. {The depth resolution of the BEV grid in the figure has been altered for visualization}. This redundancy can be leveraged to enhance the robustness and accuracy of the BEV representation, as features from multiple viewpoints provide complementary information about the scene. Therefore, in the fisheye BEV grid, the challenge lies in handling distortion and effectively integrating and leveraging the overlapping features to construct a comprehensive representation of the environment. The architectural changes needed for Fisheye cameras are explained in Section \ref{sec:arch}. \par
\subsection{Distortion-aware Learnable Pooling}
\label{sec:learnBEVPool}
The BEV features from each camera sensor are pooled into a single BEV grid. Most current methods use the \textit{BEV Pooling}~\cite{Liu2022BEVFusionMM} strategy to associate the camera BEV features to the BEV grid along with a symmetric function (e.g., mean, max, and sum) to aggregate the features within each BEV grid. However, the features in overlapping regions are treated equally and do not contextualize the sensor parameters. Our experiments showcased that it does not translate well to surround-view fisheye cameras because of the non-linear scaling and larger FOV. Also, symmetric aggregation is disadvantageous when the surround fisheye cameras do not share the same sensor configuration.

Hence, we experiment with several weighting and embedding strategies while pooling the BEV features into a single BEV grid. Specifically, we convert the pooling function into a learnable weighting function. The learnable pooling:
\begin{enumerate}
    \item primarily learns and focuses on the features relevant for an accurate BEV grid generation and semantic segmentation.
    \item plays to the strengths of the camera sensor (within the FOV), either implicitly or explicitly, while aggregating the camera BEV features.
    \item makes the final BEV grid resilient to differences in camera BEV features, especially in cases of \textit{object boundaries} and \textit{small objects}.  
\end{enumerate}
In the following sections, we discuss the different BEV pooling strategies employed. \\

\textbf{\textit{Weighted Sum}}: In the first approach, represented in Equation \ref{eqn:simbevweight}, we add a weighting parameter when summing the BEV features. The summing weights (\(W_i \in R^{C \times X \times Y}\)) are a learnable parameter for the BEV feature (\(F_{bev}^{i} \in R^{C \times X \times Y}\)) from each camera sensor, that is learned through the training process. 
\begin{equation}
\label{eqn:simbevweight}
    F_{bev}^{total} = \sum\limits_{i = 1}^N {W_i \cdot F_{bev}^{i}} 
\end{equation}

\textbf{\textit{Sensors per BEV cell}}: Secondly, we assign a set of learnable parameters for each cell in the BEV grid. Specifically, we assign one parameter for each camera contributing to the cell. These parameters weigh the features captured by each camera within the cell. The learnable parameters are initialized as one divided by the number of cameras projecting to the cell. This ensures that the initial weights are evenly distributed among the contributing cameras. If (\(w_{i, j}\)) is the learnable weight for a single BEV grid cell, and (\(N_{i, j}\)) represents the number of cameras contributing to BEV grid cell features, the weighting parameter is defined in Equation \ref{eqn:percellbevweight}.
\begin{equation}
\label{eqn:percellbevweight}
    w_{i, j} = \frac{1}{N_{i, j}}
\end{equation}

\textbf{\textit{Camera intrinsic based embedding}}: Finally, we also experiment with another pooling approach, shown in Equation \ref{eqn:lrnbevweight}, wherein we add a learnable embedding \(E_i \in R^{C \times X \times Y}\) per sensor while incorporating the sensor's intrinsic parameters \(I_K^i \in R^{3 \times 3}\). The BEV features from individual camera sensors are also scaled with their mean \(\mu_i\). We found that using a learnable embedding in pooling the BEV features focuses the network, specifically in our occlusion estimation module, thereby improving our overall results. The results from our experiments on BEV pooling are discussed in detail in Section \ref{sec:bevpoolabl}.
\begin{equation}
\label{eqn:lrnbevweight}
    F_{bev}^{total} = \sum\limits_{k = 1}^N {I_k \cdot (F_{bev}^{k}} - \mu_k) + E_k 
\end{equation}
\subsection{Occlusion reasoning in BEV Segmentation}
\label{sec:methodocc}
BEV Segmentation task is designed to represent image information in BEV space, focusing solely on grid cells perceivable by vehicle cameras. Occluded areas should not contribute to predictions to avoid speculative inference. This occlusion reasoning is particularly crucial for near-field sensing in urban driving and parking scenarios, where occlusion is prevalent.
To address this, we propose a novel multi-task approach, which includes \textit{Occlusion reasoning}, wherein the model outputs an occlusion probability for each grid cell, a feature included in our dataset. In synthetic engines, the BEV camera records the class of the closest object, whereas the desired semantic BEV labels must contain the class relevant for the ego vehicle to navigate the world. This contrast leads to the misrepresentation of BEV semantic classes when trees or bridges obstruct the BEV camera. On the contrary, there will also be label mismatches when objects are occluded in the ego vehicle's camera sensors but not in the BEV camera.

\begin{figure*}[!ht]
  \captionsetup{belowskip=0pt, font= small, singlelinecheck=false}
    \centering
    \includegraphics[width=\textwidth]{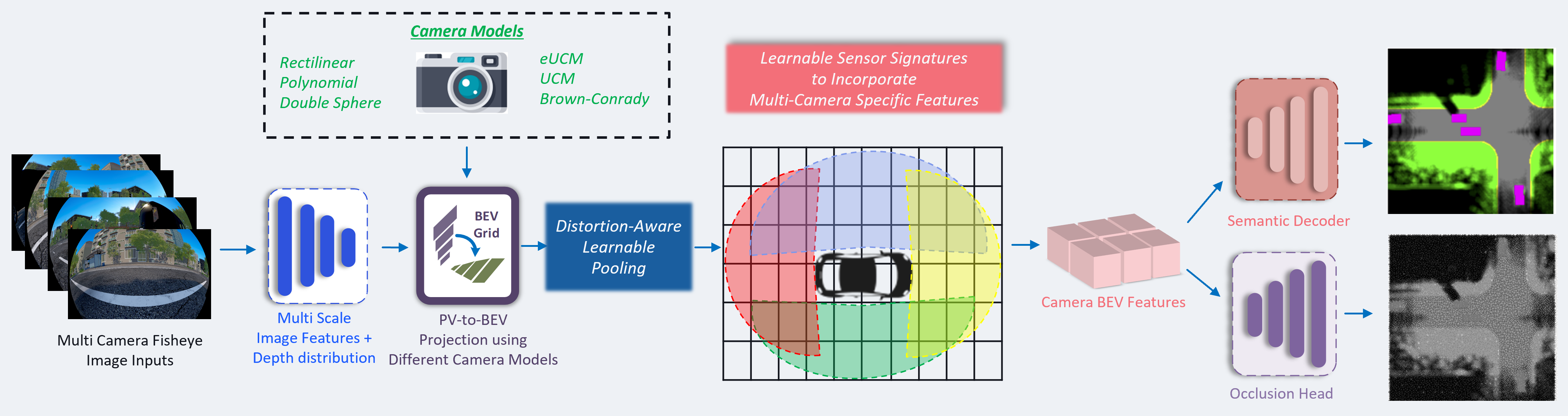}
    \caption{The network architecture converts images into image features, transforms the features into BEV space using the calibration data, and then calculates the semantic output in BEV space.}
    \label{fig:architecture}
\end{figure*}

To generate the occlusion map $p(o)$, a circular kernel is applied to the camera BEV features to count local pixel occurrences within grid cells. The occurrences are normalized with a threshold (\(\tau\)) to derive an occlusion score $p(o)$. When vehicles are visible in multiple cameras, the entire vehicle is designated as occupied and non-occluded in the BEV grid. This allows the network to comprehend vehicle dimensions, including occlusion effects. Figure~\ref{fig:architecture} visualizes occluded areas as black regions. In practice, the occupancy probability ($p^{'}(o) = 1 - p(o)$) is used for ease of loss assignment and visualization.

\subsection{Fisheye BEV Segmentation}
\label{sec:methodbevseg}

Fisheye cameras are most desirable for near-field perception because of their large FOV. Hence, we adapt the BEV segmentation to have a smaller resolution and range. For the same reason, we incorporate a multi-task head that makes classification and occupancy predictions for each cell. However, since heads operate on BEV features, we keep the model architecture of heads unchanged from \sota ~BEV segmentation techniques \cite{Liu2022BEVFusionMM}. As illustrated in Equation \ref{eqn:semsegloss}, we use weighted cross-entropy loss, based on \cite{zhang2018generalized}, for training the classification head. The occlusion head, depicted in Equation \ref{eqn:occloss}, is trained using binary cross-entropy loss. 
\begin{equation} \label{eqn:semsegloss}
    \pounds_{semantic}(p, q) = -p(o) \sum_{c_i \in C} \alpha_{i} \cdot p(c_i) \cdot \log q(c_i)
\end{equation}
The classification loss is a function of occupancy probability (\(p(o)\)). The predicted (\(p(c_i)\)) and ground truth (\(q(c_i\)) class-probabilities are weighted by \(\alpha_i\). The chosen weighting strategy for training is explained in Section \ref{sec:ablclsweight}. 
\begin{equation} \label{eqn:occloss}
    \pounds_{occ}(p(o), p(\hat{o})) = -[p(o) \cdot \log{p(\hat{o})} + (1 - p(o) \cdot \log{1 - p(\hat{o})}]
\end{equation}
Where \(p(o)\) is the occupancy probability predicted by the occupancy head and \(p(\hat{o}) \in \{0, 1\}\) is the ground truth occupancy label.

The combination of semantic classification loss (\(\pounds_{semantic}\)), and occupancy loss (\(\pounds_{occ}\)), presented in Equation \ref{eqn:totloss}, is used to train the BEV Segmentation model end-to-end. When combined with semantic loss, we introduce a weighting parameter (\(\lambda\)) to occupancy oss. This focuses the training on the BEV Segmentation task and helps in training stability and convergence.
\begin{equation} \label{eqn:totloss}
    \pounds_{total} = \pounds_{semantic} + \lambda \cdot \pounds_{occ}
\end{equation}

\section{Implementation}
\label{sec:implementation}

\subsection{Datasets}


\begin{figure}
    \centering
    \captionsetup{belowskip=0pt, font= small, singlelinecheck=false}
    \includegraphics[width=1\linewidth]{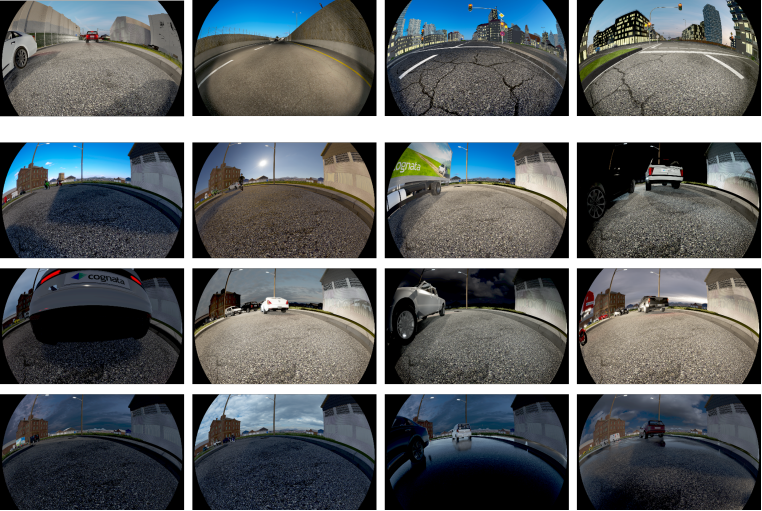}
    \caption{\textbf{Row 1}: Four different scenes from the Cognata training set. \textbf{Rows 2-4}: 12 different weather and traffic situations from the fifth training scene}
    \label{fig:enter-label}
\end{figure}

For fisheye BEV segmentation, there are no real-world datasets, but there are two possible synthetic datasets, namely SynWoodScape\cite{sekkat2022synwoodscape} and F2BEV (FB-SSEM dataset)~\cite{samani2023f2bev}. These datasets do not address the occlusion problem described in the previous Occlusion reasoning section. Besides this issue, F2BEV is confined to simulating just parking lots and is not suitable for general driving. SynWoodScape has released only a small initial sample of the dataset.
Therefore, we generated a synthetic dataset using the Cognata simulation platform. To overcome the occlusion problem, we enrich the information from the BEV camera using additional sensor information from other vehicle cameras. We project the perspective semantic labels from other camera sensors in the vehicle to 3D space to rectify the BEV camera semantic labels. Our final dataset comprises four fisheye cameras and six pinhole cameras with \(1920\times1208\) resolution, BEV ground truth images (\(400 \times 400\)) with five semantic classes: \textit{invalid, vehicles, lane markings, street, and background}.

In addition, we generate an occlusion probability map indicating the likelihood of occlusion in each grid cell.  
The training dataset comprises five virtual scenes of about 90 seconds each from Pittsburgh, Munich, and HaSharon. Each scene is simulated with different traffic, weather, and daylight conditions. The whole dataset contains more than 12,000 different frames corresponding to 50,000 fisheye images. Sample images from the dataset can be seen in Figure \ref{fig:enter-label}. \par

\begin{table}[t]
\captionsetup{belowskip=-15pt, font= small, singlelinecheck=false}
\caption{\textbf{Quantitative Performance of~\algoname} on different test dataset slices. Input image size is \(480 \times 302\).}
\label{tab:fullmetrics}
\centering{
\begin{adjustbox}{width=\columnwidth}
\begin{tabular}{c|c|c|cccccc}
\toprule
\multirow{2}{*}{\textit{Method}} & \multirow{2}{*}{\textit{Image Type}} & \multirow{2}{*}{\textit{Dataset}}
& \multicolumn{6}{c}{\cellcolor[HTML]{a5eb8d}\textit{IoU}} \\ \cline{4-9} & & & 
\multicolumn{1}{c|}{\cellcolor[HTML]{00b050}\textit{mIoU}} 
& \cellcolor[HTML]{fb9a99}\textit{Occlusions}
& \cellcolor[HTML]{96bbce}\textit{Vehicles}          
& \cellcolor[HTML]{fc4538}\textit{Markings}          
& \cellcolor[HTML]{fdbf6f}\textit{Street}            
& \cellcolor[HTML]{ab9ac0}\textit{Background} \\ \hline
\multirow{3}{*}{\textit{LSS~\cite{Philion2020LiftSS}}} 
& \multirow{3}{*}{\begin{tabular}[c]{@{}c@{}}Cyl.\\ Rectified\end{tabular}} 
& \cellcolor{gray95}Easy & \multicolumn{1}{c|}{\cellcolor{gray95}0.659} & \multicolumn{1}{c}{\cellcolor{gray95}-} & \multicolumn{1}{c}{\cellcolor{gray95}0.607} & \multicolumn{1}{c}{\cellcolor{gray95}0.379} & 
\multicolumn{1}{c}{\cellcolor{gray95}0.773} & \cellcolor{gray95}0.875 \\ 

& & \cellcolor{gray9}Medium & \multicolumn{1}{c|}{\cellcolor{gray9}0.540} & \multicolumn{1}{c}{\cellcolor{gray9}-} & \multicolumn{1}{c}{\cellcolor{gray9}0.539} 
& \multicolumn{1}{c}{\cellcolor{gray9}0.214} & \multicolumn{1}{c}{\cellcolor{gray9}0.645} & \cellcolor{gray9}0.763 \\

& & \cellcolor{gray85}Hard & \multicolumn{1}{c|}{\cellcolor{gray85}0.314} & \multicolumn{1}{c}{\cellcolor{gray85}-} & \multicolumn{1}{c}{\cellcolor{gray85}0.317} 
& \multicolumn{1}{c}{\cellcolor{gray85}0.114} & \multicolumn{1}{c}{\cellcolor{gray85}0.523} & \cellcolor{gray85}0.303 \\ 

\midrule

\multirow{3}{*}{\algoname} & \multirow{3}{*}{\begin{tabular}[c]{@{}c@{}}Raw\\ Distorted\end{tabular}} 
& \cellcolor{gray95}Easy & \multicolumn{1}{c|}{\cellcolor{gray95}0.796} & \multicolumn{1}{c}{\cellcolor{gray95}0.815} & \multicolumn{1}{c}{\cellcolor{gray95}0.776} 
& \multicolumn{1}{c}{\cellcolor{gray95}0.517} & \multicolumn{1}{c}{\cellcolor{gray95}0.895} & \cellcolor{gray95}0.978 \\

& & \cellcolor{gray9}Medium & \multicolumn{1}{c|}{\cellcolor{gray9}0.690} & \multicolumn{1}{c}{\cellcolor{gray9}0.682} & \multicolumn{1}{c}{\cellcolor{gray9}0.764} 
& \multicolumn{1}{c}{\cellcolor{gray9}0.364} & \multicolumn{1}{c}{\cellcolor{gray9}0.858} & \cellcolor{gray9}0.782 \\

& & \cellcolor{gray85}Hard & \multicolumn{1}{c|}{\cellcolor{gray85}0.466} & \multicolumn{1}{c}{\cellcolor{gray85}0.666} & \multicolumn{1}{c}{\cellcolor{gray85}0.464} 
& \multicolumn{1}{c}{\cellcolor{gray85}0.176} & \multicolumn{1}{c}{\cellcolor{gray85}0.572} & \cellcolor{gray85}0.449 \\
\bottomrule
\end{tabular}
\end{adjustbox}
}
\end{table}
\begin{table}[t]
\caption{\textbf{Ablation Study of~\algoname~on \textit{Medium} test dataset}. \footnotesize Under \textit{Class Weight} ablation, \textit{\textbf{X-X-X-X}} represents weights for Vehicles, Lane Markings, Street and Background . }
\label{tab:clsweight}
\centering{
\begin{adjustbox}{width=\columnwidth}
\begin{tabular}{@{}lllllll@{}}
\toprule
\multicolumn{1}{l|}{\cellcolor[HTML]{96bbce}\textit{Approaches}} 
& \multicolumn{6}{c}{\cellcolor[HTML]{a5eb8d}\textit{IoU}} \\ 
\cmidrule(l){2-7} 
\multicolumn{1}{l|}{\multirow{-2}{*}{}} 
& \multicolumn{1}{l|}{\cellcolor[HTML]{00b050}\textit{mIoU}} 
& \cellcolor[HTML]{fb9a99}\textit{Occlusion} 
& \cellcolor[HTML]{96bbce}\textit{Vehicles} 
& \cellcolor[HTML]{fc4538}\textit{Markings} 
& \cellcolor[HTML]{fdbf6f}\textit{Street} 
& \cellcolor[HTML]{ab9ac0}\textit{Background} \\ 
\midrule
\multicolumn{7}{c}{\cellcolor[HTML]{FFCCC9}\textit{\textbf{Class Weights}}} \\ 
\midrule

\multicolumn{1}{c|}{13-3-1-1} & \multicolumn{1}{l|}{0.658} & \textbf{0.687} & 0.716 & 0.346 & 0.803 & 0.736 \\
\multicolumn{1}{c|}{1-1-1-1} & \multicolumn{1}{l|}{\textbf{0.690}} & 0.682 & \textbf{0.764} & \textbf{0.364} & \textbf{0.858} & \textbf{0.782} \\
\midrule

\multicolumn{7}{c}{\cellcolor[HTML]{FFCE93}\textit{\textbf{Loss for Occluded areas}}} \\ 
\midrule
\multicolumn{1}{c|}{\ch} & \multicolumn{1}{l|}{\textbf{0.658}} & \textbf{0.687} & 0.716 & \textbf{0.346} & \textbf{0.803} & \textbf{0.736} \\
\multicolumn{1}{c|}{\xm} & \multicolumn{1}{l|}{0.631} & 0.675 & \textbf{0.724} & 0.342 & 0.752 & 0.662 \\

\midrule

\multicolumn{7}{c}{\cellcolor[HTML]{7d9ebf}\textit{\textbf{BEV Pooling methods}}} \\
\multicolumn{1}{c|}{Sum}         & \multicolumn{1}{l|}{0.666}   & 0.665      & 0.719   & 0.308             & \textbf{0.862}  & 0.776    \\
\multicolumn{1}{c|}{Maxpool}      & \multicolumn{1}{l|}{0.656}   & 0.602               & 0.748             & 0.309             & 0.857           & 0.764               \\
\multicolumn{1}{l|}{Weighted Sum (ours)}                 & \multicolumn{1}{l|}{0.667}          & 0.607               & 0.755    & 0.325             & 0.860           & 0.769               \\
\multicolumn{1}{l|}{Sensors per BEV Cell (ours)}         & \multicolumn{1}{l|}{0.679}          & 0.644               & 0.759             & 0.363             & 0.861           & 0.774               \\
\multicolumn{1}{c|}{\makecell{Cam Intrin + Emb (ours)}} & \multicolumn{1}{l|}{\textbf{0.690}} & \textbf{0.682}      & \textbf{0.764}             & \textbf{0.364}    & 0.858           & \textbf{0.782} \\
\bottomrule
\end{tabular}
\end{adjustbox}
}
\end{table}

\subsection{Architecture Details}
\label{sec:arch}
As described in Section \ref{sec:method}, we select the architectural approach from Lift-Splat-Shoot~\cite{Philion2020LiftSS} to implement the BEV segmentation using fisheye cameras. Given the near-field applications of fisheye cameras, we choose a BEV grid cell size of \(0.25m\) instead of \(0.5m\), which is prevalent in pinhole camera approaches. Further, we set a coverage range of \(25m\) from the ego vehicle longitudinally and latitudinally instead of the commonly used \(50m\) range.

Also, we analyzed the architecture for accuracy bottlenecks and found it to be input resolution. Hence we increased it to \(480 \times 302\), from \(352 \times 128\), almost doubling the number of input pixels. Given the larger FOV of the fisheye images, this also follows our intuitive reasoning. Finally, to address the relative imbalance in the dataset volume between open-source pinhole datasets and our simulated Cognata dataset, we test several combinations of dropout layers in the network to prevent overfitting. Additionally, we incorporate color jittering augmentation and other commonly used data augmentation techniques such as scaling, rotation, and cropping. We use a contrast variation of \(0.5\), saturation and brightness variation of \(0.3\) each, and a hue variation of \(0.05\).
\section{Experiments and Results}

We perform several training runs to evaluate implementation details and improve the results. We use a GPU server with four Nvidia Quadro RTX5000 GPUs with a batch size of 64. We use Adam optimizer \cite{kingma2014adam} with a learning rate of \(10^{-4}\). All implementation is done in PyTorch \cite{paszke2019pytorch}. 

We generate a test set of three scenes: \textit{easy, medium, and hard}. The easy scene is a traffic and weather variation of a scene that's part of the training set. Therefore, the road geometry is familiar to the model in an easy scene, while the vehicle configuration is still unknown. The medium scene is a previously unseen highway scene, and the hard scene is a previously unseen urban scene with complex road geometry. We use Intersection over Union (IoU), defined in Equation \ref{eqn:iou}, as the primary metric for semantic classification. It is computed with predicted semantic class (\(y\)) and ground truth semantic class (\(\hat{y}\) ). We track and report the IoU for each class separately to account for the skew in classes (background vs pedestrians/vehicles). We also use IoU to predict occupancy by converting it to binary masks.\par
\begin{equation} \label{eqn:iou}
    IoU(y, \hat{y}) = \frac{|y \cap \hat{y}|}{|y \cup \hat{y}|}
\end{equation}
Finally, we report the mIoU, which is typically just the mean of all IoU scores of the classification head. As we want a single metric to compare the performance between different networks, we want to incorporate the score for occlusion into the mIoU score. Therefore, the mIoU score we report in the following is the average from the five IoU scores for occlusion, vehicles, lane markings, street, and background. Table \ref{tab:fullmetrics} shows the results of training and test datasets. Naturally, the training performance scores highest. Comparing the three test sets shows the performance drop with the increasing difficulty level. The medium test set scores are significantly lower than the easy test set, especially for the lane marking and the background. However, the network learns to position the vehicles quite well.

 We also compared against standard baseline~\cite{Philion2020LiftSS} by applying a cylindrical projection to rectify the fisheye images. Cylindrical projection preserves the vertical lines from fisheye images. The results in Table \ref{tab:fullmetrics} indicate that our model performs better than the rectification process. It is an intuitive outcome, given that rectification introduces unnecessary object stretching and interpolation artifacts to the image in addition to losing some of the field of view.
\subsection{Ablation Studies}
\subsubsection{Class Weighting}
\label{sec:ablclsweight}

As discussed in Section \ref{sec:methodbevseg}, we apply a non-uniform class weighting to improve the
overall performance by focusing more on underrepresented classes. During the experiment, we used a class weighting of (\(13 \times, 3 \times, 1 \times, 1 \times\)) for vehicles, lane markings, streets, and backgrounds, respectively.
We analyze the effect of the class-weighted loss and compare the configuration mentioned above with a uniformly weighted loss. The model with uniformly weighted loss performs better for almost all classes. The overall performance of the model with uniform class loss is \(0.032\) mIoU points better than the performance for the model with weighted loss, as stated in Table \ref{tab:clsweight}.\par

Our explanation for this counter-intuitive behavior is that it is a different task than image classification. Most literature about dataset and loss balancing studies the problem of image classification. There, the classes are often equally challenging, and improving the performance of underrepresented classes makes sense. For semantic BEV segmentation, though, not only the occurrence of a class at the output of the network matters. Additionally, the presence and detectability of a class in the input domain are essential.\par
\subsubsection{Occupancy Prediction}

Given the geometry-based approach for Camera to BEV space, we hypothesized that the network cannot estimate areas occluded in the image well. Therefore, the network guesses occluded regions based on image statistics, deteriorating the overall quality. We apply an occlusion-based loss function to circumvent this guessing or hallucinating on occluded areas. When regions are occluded, we do not assign a loss to them, so the network is not encouraged to hallucinate. Loss removal for occluded regions avoids the network updating its weights based on these occluded regions. Further, we compute the IoU only for grid cells that are at least \(50\%\) visible so that our metrics indicate the performance on perceived regions and not
on hallucinated ones. Table \ref{tab:clsweight} illustrates the impact of occupancy loss on the segmentation module.

\subsubsection{Comparing different BEV Pooling methods}
\label{sec:bevpoolabl}
Incorporating learnable parameters for each cell in the BEV grid enables adaptive feature pooling, allowing us to consider each sensor's contribution to the BEV grid. This results in a more context-aware representation of features. This adaptivity enhances the ability to capture relevant information from different cameras and their varying characteristics. Further, it enables the pooling operation to leverage the strengths of various sensors, considering factors such as their field of view, resolution, or proximity to the target object. Specifically, for the BEV segmentation task, the finer cell-level adaptivity of BEV pooling yields a better semantic boundary in the case of small objects. It is evidenced in our results in Table \ref{tab:clsweight}. We see much better results from our pooling method that factor in sensor intrinsics than other naive pooling approaches. Using intrinsics and embedding for BEV pooling also improves our occlusion IoU significantly, further strengthening our claims. 
\begin{figure}[!t]
\captionsetup{belowskip=0pt, font= small, singlelinecheck=false}
  \includegraphics[width=\linewidth]{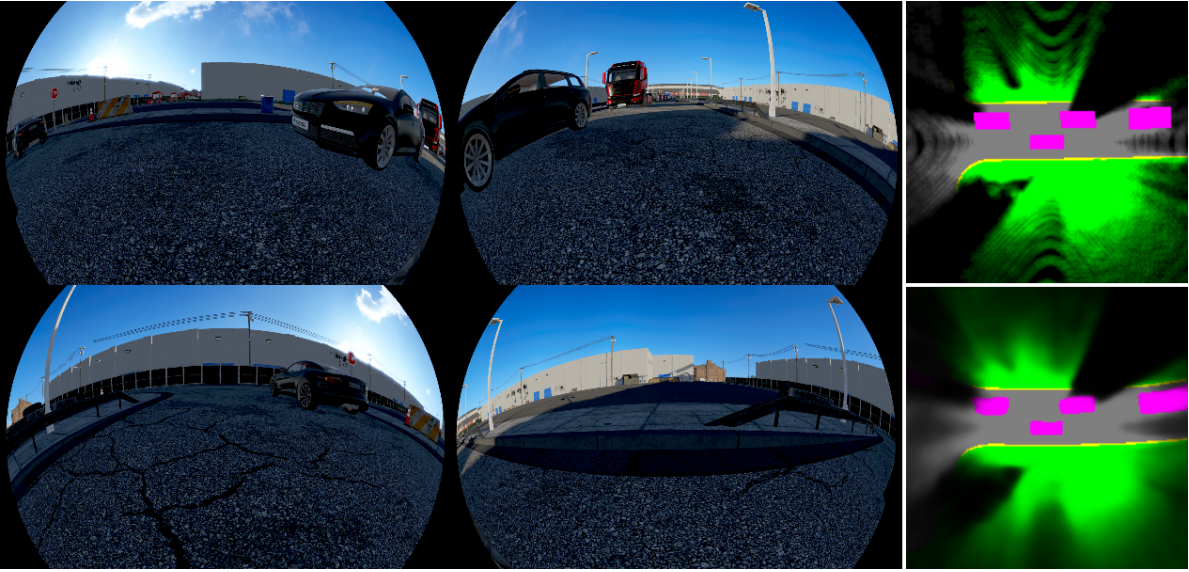}
  \caption{\algoname~ results on \textbf{Easy} scene. \textit{Left}: The 4x surround camera images. \textit{Top right}: The ground truth. \textit{Bottom right}: Model Prediction.}
  \label{fig:quali_res_easy}
\end{figure}
\begin{figure}[!ht]
\captionsetup{belowskip=0pt, font= small, singlelinecheck=false}
  \includegraphics[width=\linewidth]{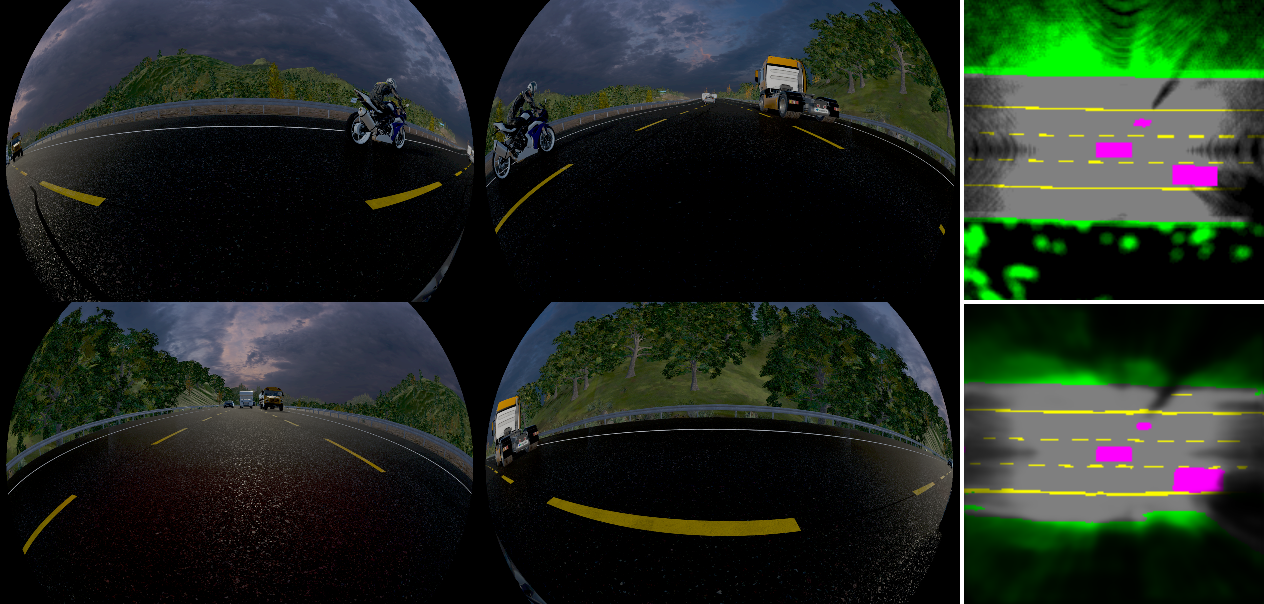}
  \caption{\algoname~ results on \textbf{Medium} scene. \textit{Left}: The 4x surround camera images. \textit{Top right}: The ground truth. \textit{Bottom right}: Model Prediction.}
  \label{fig:quali_res_med}
\end{figure}
\begin{figure}[!ht]
\captionsetup{belowskip=0pt, font= small, singlelinecheck=false}
  \includegraphics[width=\linewidth]{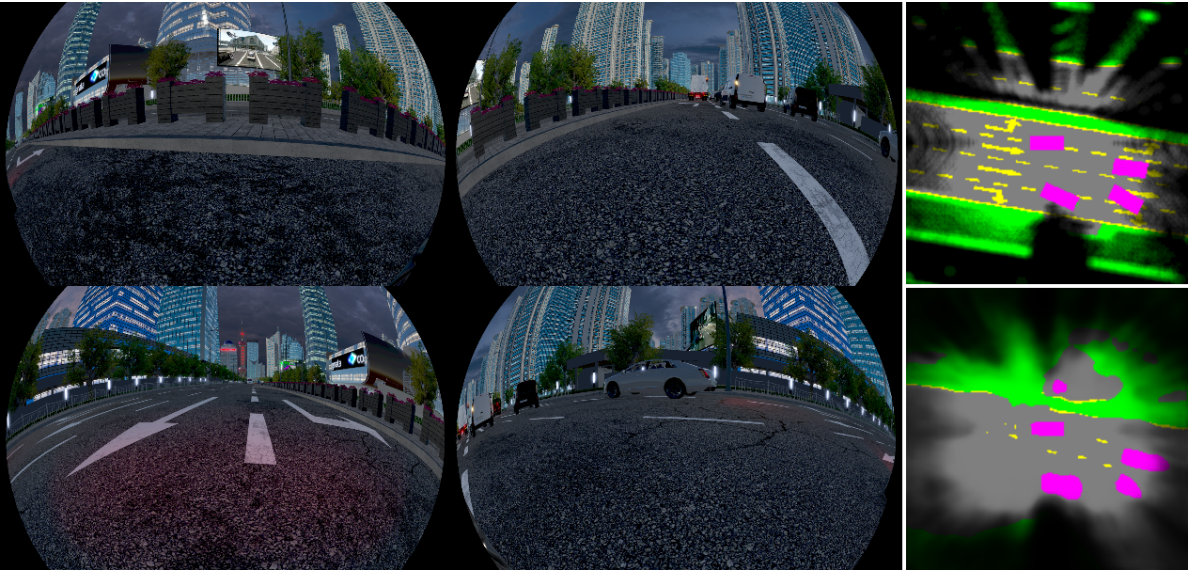}
  \caption{\algoname~ results on \textbf{Hard} scene. \textit{Left}: The 4x surround camera images. \textit{Top right}: The ground truth. \textit{Bottom right}: Model Prediction.}
  \label{fig:quali_res_hard}
\end{figure}
\subsection{Qualitative Results}

We provide qualitative results of~\algoname~ on the test data. Figure \ref{fig:quali_res_easy} shows the 4x fisheye images, ground truth, and prediction results for an easy scene. We provide complete qualitative results of~\algoname~ on the test data at \url{https://streamable.com/ge4v51}. As a similar scene is part of the training data, the network outputs an accurate semantic BEV segmentation. The network performs an excellent fusion of the Fisheye images into a single BEV image. 

In case of \textit{medium} test scene, shown in Figure \ref{fig:quali_res_med}, the segmentation complexity can seen in road class beyond 10m. However, vehicles and lane markings are still detected accurately. 
On a hard test scene, as shown in Figure \ref{fig:quali_res_hard},~\algoname~is challenged the most, with complicated lane markings. This is intuitive considering out-of-distribution data is always challenging for supervised methods.

\section{Conclusion}
\label{sec:conclusion}

In this work, we introduce a novel pipeline to perform semantic Bird's Eye View (BEV) segmentation using images from fisheye cameras. We perform our experiments on a synthetic dataset due to the lack of public datasets with fisheye cameras. Future directions for our work include bridging the reality gap between synthetic and real-world data and extending to additional tasks with fisheye cameras. We hope that this work encourages further research in performing BEV perception for fisheye cameras.


\section*{Acknowledgment}

The synthetic data used in this paper were generated using a commercial synthetic engine \url{www.cognata.com}. The authors thank our colleague Anamarija Fofonjka for the detailed review and feedback.
\bibliographystyle{IEEEtran}
\bibliography{ieee}
\clearpage
\end{document}